\documentclass{article}
\usepackage{times}  
\usepackage{helvet}  
\usepackage{courier}  
\usepackage[hyphens]{url}  
\usepackage{graphicx} 
\usepackage{caption} 
\usepackage{spconf,amsmath,graphicx,hyperref}
\usepackage{algorithm}
\usepackage{algorithmic}
\usepackage{color}
\usepackage{xcolor}
\usepackage{newfloat}
\usepackage{listings}
\usepackage{bibentry}
\usepackage{amsmath}
\usepackage{arydshln}
\usepackage{amsfonts}
\usepackage{enumitem}
\usepackage{booktabs}
\usepackage{multirow}
\usepackage{amssymb}
\usepackage{subfigure}
\usepackage{subcaption}

\title{Legal$\Delta$: Enhancing Legal Reasoning in LLMs via Reinforcement Learning with Chain-of-Thought Guided Information Gain}
\newcommand{\equalcontrib}{\textsuperscript{*}}
\name{    
    Xin Dai\textsuperscript{\rm 1\equalcontrib},
    Buqiang Xu\textsuperscript{\rm 1\equalcontrib}\thanks{\textsuperscript{*} Equal contribution.},
    Zhenghao Liu\textsuperscript{\rm 1$\dagger$},
    Yukun Yan\textsuperscript{\rm 2},
    Huiyuan Xie\textsuperscript{\rm 2},
    Xiaoyuan Yi\textsuperscript{\rm 3},
    Shuo Wang\textsuperscript{\rm 2},
    Ge Yu\textsuperscript{\rm 1}
    \thanks{$\dagger$ Corresponding author.}}
\address{
    \textsuperscript{\rm 1}School of Computer Science and Engineering, Northeastern University, Shenyang, China\\
    \textsuperscript{\rm 2}Department of Computer Science and Technology, Institute for AI, Tsinghua University, Beijing, China\\
    \textsuperscript{\rm 3}Microsoft Research Asia, Beijing, China\\
}
\begin{document}
\ninept
\maketitle
\begin{abstract}
Legal Artificial Intelligence (LegalAI) has achieved notable advances in automating judicial decision-making with the support of Large Language Models (LLMs). However, existing legal LLMs still struggle to generate reliable and interpretable reasoning processes. They often default to fast-thinking behavior by producing direct answers without explicit multi-step reasoning, limiting their effectiveness in complex legal scenarios that demand rigorous justification.
To address this challenge, we propose Legal$\Delta$, a reinforcement learning framework designed to enhance legal reasoning through \emph{chain-of-thought guided information gain}. During training, Legal$\Delta$ employs a dual-mode input setup—comprising direct answer and reasoning-augmented modes—and maximizes the information gain between them. This encourages the model to acquire meaningful reasoning patterns rather than generating superficial or redundant explanations.
Experimental results on multiple legal reasoning tasks demonstrate that Legal$\Delta$ outperforms strong baselines and consistently produces more robust and trustworthy legal judgments without relying on labeled preference data. All code and data are available at https://github.com/NEUIR/LegalDelta.
\end{abstract}
\section{Introduction}
\label{sec:intro}

The rapid development of LegalAI systems—referring to artificial intelligence technologies applied to legal practice and judicial processes—is fundamentally reshaping the legal landscape. By automating complex tasks such as legal information extraction~\cite{rabelo2022overview}, intelligent case retrieval~\cite{zhang2023case}, and judgment prediction~\cite{10.1007/978-981-95-3453-1_23}, LegalAI has significantly improved the efficiency and accessibility of legal services, demonstrating substantial value in real-world applications such as contract analysis, legal dispute resolution, and regulatory compliance. Building on these advances, recent research has increasingly explored Large Language Models (LLMs) as a foundation for developing more effective legal assistants.



Recently, researchers have increasingly adopted Reinforcement Learning (RL) frameworks such as Group Relative Policy Optimization (GRPO), which enables models to learn from feedback and optimize their behavior through trial-and-error. Large Reasoning Models (LRMs) like QwQ~\cite{team2025qwq} and DeepSeek-R1~\cite{guo2025deepseek} have demonstrated strong reasoning capabilities across specialized domains including medicine~\cite{sun2025reasonmed}, mathematics~\cite{ahn2024large}, and programming~\cite{el2025competitive}. Recent work has extended these ideas to the legal domain by distilling reasoning traces from powerful RL-trained models such as DeepSeek-V3~\cite{liu2024deepseek}, as in LawGPT. These methods typically rely on rule-based rewards in Reinforcement Learning with Verifiable Rewards (RLVR)~\cite{mroueh2025reinforcement}, which primarily verify the correctness of outputs to optimize LRMs. However, the absence of more nuanced verifiable reward modeling limits their effectiveness in optimizing legal LLMs, as RLVR often overlooks the quality and depth of the reasoning process. Consequently, how to effectively optimize Chain-of-Thought (CoT) within the RL paradigm to achieve more convincing legal reasoning remains an open question.

To address these challenges, this paper presents Legal$\Delta$, a reinforcement learning framework designed to improve legal reasoning in LLMs by incorporating \textbf{Chain-of-Thought guided Information Gain}. Built upon the GRPO paradigm, Legal$\Delta$ introduces an information-gain based reward mechanism into the RLVR setting, encouraging LLMs to explore diverse reasoning trajectories and develop robust reasoning strategies.
Specifically, Legal$\Delta$ quantifies information gain by leveraging the mathematical equivalence between model logits and Q-values in reinforcement learning. By comparing model responses under direct inference mode and chain-of-thought prompting mode, it computes confidence differentials that capture both (i) pointwise mutual information between reasoning steps and final answer correctness, and (ii) global confidence shifts indicating whether reasoning improves decision reliability. This reward is then used to guide policy optimization, making Legal LLMs discover and reinforce high-quality reasoning behaviors that balance prediction accuracy and confidence.

Our experiments on various LLMs demonstrate the effectiveness of Legal$\Delta$, which outperforms all baseline models and achieves approximately 10\% improvement across multiple legal reasoning tasks. Notably, Legal$\Delta$ maintains its advantages on out-of-domain tasks, demonstrating strong generalization capability.
Further analysis reveals that Legal$\Delta$ effectively enhances model confidence through information gain-enhanced reward modeling, which progressively improves logits, increases the prominence of legal tokens, and further enhances reasoning quality.

\section{Methodology}
\label{sec:pagestyle}
In this section, we introduce the reinforcement learning based legal optimization method (Section~\ref{Training Strategy}) and the information-gain-enhanced reward mechanism (Section~\ref{Information-Gain Enhanced Reward Mechanism}).

\begin{figure}[t]
    \centering
    \includegraphics[width=\columnwidth]{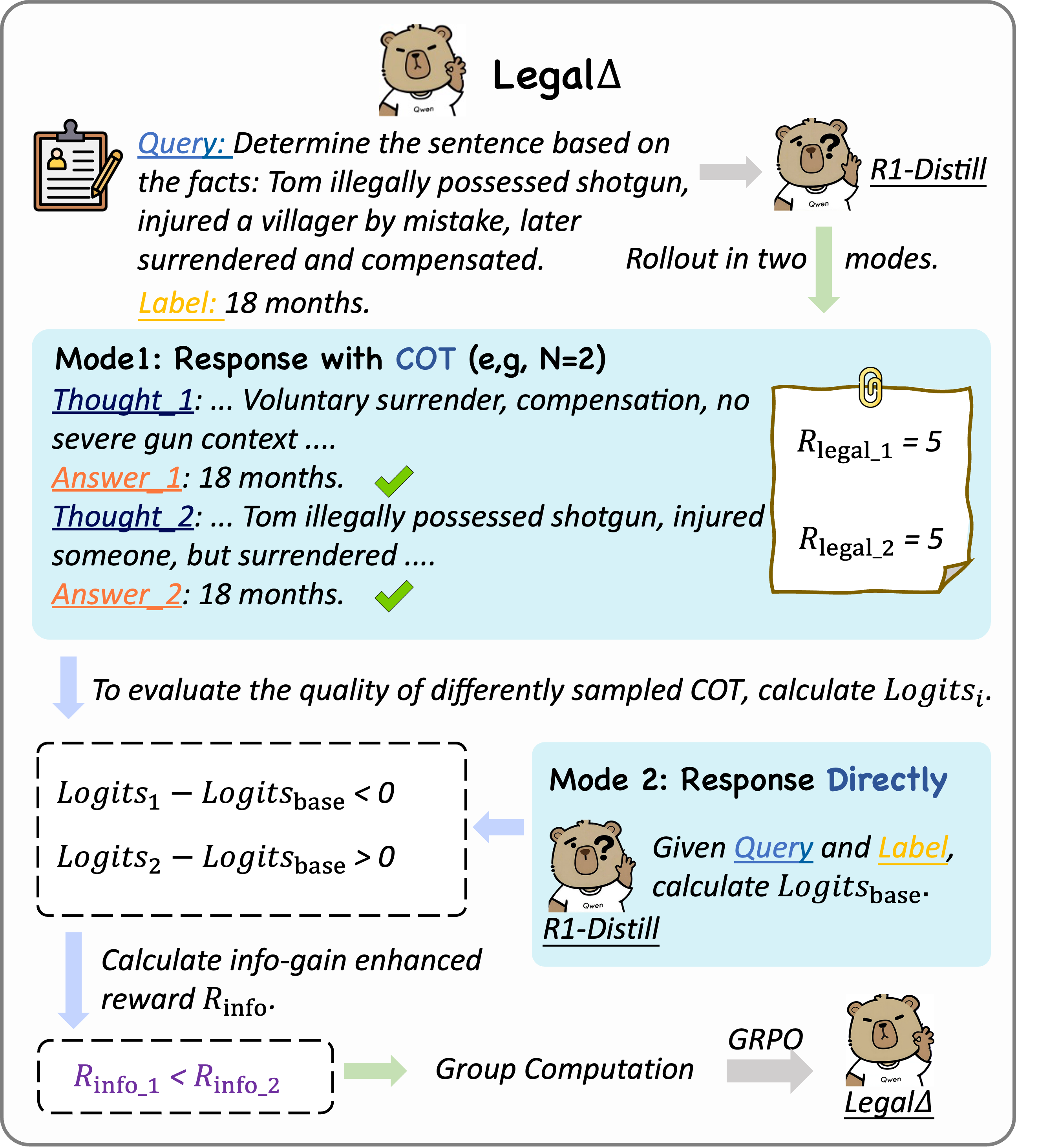}
    \caption{Illustration of Legal$\Delta$.}
    \label{fig:main_photo}
\end{figure}
\subsection{Enhancing the Reasoning Capability of LLMs via Reinforcement Learning}
Given a query $q$, our model $\mathcal{M}_\text{Legal}$ produces a structured output:
\begin{equation}
    q \rightsquigarrow \mathcal{M}_\text{Legal} \rightsquigarrow o = (r, a),
\end{equation}
where $r$ denotes the step-by-step reasoning process and $a$ is the final answer. This structured output ensures both interpretability and reliability, which are crucial in legal applications.

General-domain LLMs often fall short in domain-specific reasoning tasks. Although Supervised Fine-Tuning (SFT)~\cite{gururangan2020don} can adapt models to target domains, it also risks overfitting and catastrophic forgetting~\cite{luo2025empirical}, thereby weakening general reasoning capabilities.
To address this, we propose a RL-based training framework. The first stage employs SFT to guide the LLM to imitate the expert-like reasoning behaviors. The second stage applies reinforcement learning to further optimize LLMs based on performance feedback.

\label{Training Strategy}
\textbf{Supervised Fine-Tuning.}
We construct $\mathcal{M}_\text{R1-Distill}$ via SFT using reasoning-enhanced legal data. Specifically, we leverage the high-quality reasoning chains generated by DeepSeek-R1~\cite{guo2025deepseek} to construct a dataset pairing legal queries with structured and expert-like reasoning steps.
The SFT loss is defined as the standard cross-entropy loss over the full reasoning sequence:
\begin{equation}
\begin{aligned}
    \mathcal{L}_\text{SFT} 
    &= -\mathbb{E}_{(q,r,a)\sim\mathcal{D}_\text{train}} \left[\log P_{\theta}(r, a | q)\right],
\end{aligned}
\end{equation}
where $\mathcal{D}_\text{train}$ represents the training dataset and $P_{\theta}(r,a | q)$ represents the probability of generating the complete output containing both reasoning chain and final answer given the query.

\textbf{Reinforcement Learning.}
As shown in Figure~\ref{fig:main_photo}, building upon the distilled model, we apply Group Relative Policy Optimization (GRPO)~\cite{shao2024deepseekmath} to further enhance the model's reasoning ability on these legal tasks. Unlike traditional RL approaches that require separate critic models, GRPO compares intra-group outputs to estimate advantages, making it effective for the scenarios where outputs can vary widely in quality.

Within the GRPO framework, our policy model $\pi_\theta$ generates multiple reasoning trajectories $\{o_i\}_{i=1}^G$ for each legal query. Then we can use the GRPO training method to optimize LLMs:
\begin{equation}
\begin{aligned}
J_{GRPO}(\theta) &= 
\mathbb{E}_{q\sim\mathcal{D}, \{o_i\}_{i=1}^G \sim \pi_{\theta_{old}}(\cdot|q)} \left[\frac{1}{G}\sum_{i=1}^{G} L_i(\theta)\right] \\
&\quad - \beta D_{KL}(\pi_{\theta_{old}}||\pi_{\theta}),
\end{aligned}
\end{equation}
where $L_i(\theta) = -\log \pi_\theta(o_i|q) \cdot A_i^G$, and $A_i^G$ is the group-wise advantage.
The effectiveness of GRPO critically depends on the quality of reward signals. Legal$\Delta$ defines a structured reward function as:
\begin{equation}\label{eq:reward_raw}
R(o_i) = R_{\text{format}}(o_i) + R_{\text{legal}}(a_i),
\end{equation}
where the \textbf{Format Reward ($R_{\text{format}}(o_i)$)} ensures outputs adhere to a structured format, with \texttt{<reasoning>} and \texttt{<answer>} tags separating the reasoning chain from the final answer. 
And the \textbf{Legal Reward ($R_{\text{legal}}(a_i)$)} evaluates the correctness and legality of the final answer $a_i$ using fine-grained, task-specific criteria. 
    
\begin{table*}[t]
\centering
\small  
\setlength{\arraycolsep}{6pt}  
\begin{tabular}{l@{\hspace{10pt}}c@{\hspace{5pt}}c@{\hspace{10pt}}c@{\hspace{5pt}}c@{\hspace{10pt}}c@{\hspace{10pt}}c@{\hspace{10pt}}c@{\hspace{10pt}}c@{\hspace{10pt}}c@{\hspace{10pt}}c@{\hspace{10pt}}c@{\hspace{10pt}}c@{\hspace{10pt}}c@{\hspace{10pt}}c}
\toprule
\multirow{3}{*}{\textbf{Method}} & \multicolumn{8}{c}{\textbf{In-Domain}} & \multicolumn{6}{c}{\textbf{Out-of-Domain}} \\
\cmidrule(lr){2-9} \cmidrule(lr){10-15}
& \multicolumn{2}{c}{\textbf{SPP-F}} & \multicolumn{2}{c}{\textbf{CCP}} & \textbf{SLP} & \textbf{CAS} & \textbf{CAC} & \multirow{2}{*}{\textbf{Avg.}} & \textbf{PAE} & \textbf{NJE} & \textbf{PFE} & \textbf{UNGEE} & \textbf{LBK} & \multirow{2}{*}{\textbf{Avg.}} \\
& \textbf{\textit{F1}} & \textbf{\textit{Jac}} & \textbf{\textit{F1}} & \textbf{\textit{Jac}} & \textbf{\textit{Acc}} & \textbf{\textit{Acc}} & \textbf{\textit{Acc}} & & \textbf{\textit{Acc}} & \textbf{\textit{Acc}} & \textbf{\textit{Acc}} & \textbf{\textit{Acc}} & \textbf{\textit{Acc}} & \\
\midrule
\multicolumn{15}{c}{\textit{Law Model}} \\
\hdashline
LexiLaw-6B & 13.15 & 13.14 & 39.99 & 35.63 & 8.30 & 20.80 & 35.80 & 23.83 & 23.73 & 20.11 & 31.76 & 31.56 & 40.36 & 29.50 \\
Hanfei-7B & 0.26 & 2.65 & 30.96 & 27.38 & 17.10 & 23.60 & 39.40 & 20.19 & 21.19 & 14.53 & 31.76 & 24.38 & 29.09 & 24.19 \\
Fuzi-7B & 25.19 & 27.51 & 55.93 & 49.03 & 15.60 & 0.70 & 47.20 & 31.59 & 26.27 & 24.95 & 33.53 & 28.13 & 40.73 & 30.72 \\
DiscLaw-13B & 38.75 & 48.16 & 52.87 & 48.03 & 41.10 & 0.00 & 16.80 & 35.10 & 40.68 & 42.09 & 57.06 & 50.94 & 54.91 & 49.14 \\
ChatLaw-13B & 33.28 & 32.55 & 27.90 & 24.54 & 11.60 & 28.80 & 41.40 & 28.58 & 31.36 & 27.56 & 42.35 & 35.62 & 41.09 & 35.60 \\ 
\midrule
\multicolumn{15}{c}{\textit{Qwen2.5-7B-Instruct}} \\
\hdashline
Zero-shot & 77.10 & 75.46 & 53.81 & 48.43 & 48.30 & 59.80 & 66.80 & 61.39 & 66.94 & 58.84 & 90.00 & 84.37 & 82.90 & 76.61 \\
SFT (DiscLaw) & 72.71 & 67.72 & 41.79 & 37.56 & 51.60 & 38.40 & 55.20 & 52.14 & 47.46 & 42.09 & 68.23 & 55.00 & 62.55 & 55.07 \\
DPO & 77.23 & 73.56 & 55.28 & 50.05 & 68.40 & 56.60 & 64.40 & 63.65 & 66.10 & 60.34 & 90.59 & \textbf{85.94} & 84.73 & 77.54 \\
R1-Distill & 80.14 & 78.12 & 54.73 & 49.92 & 50.20 & 55.60 & 69.20 & 62.56 & 66.10 & 62.75 & 91.17 & 82.81 & 84.72 & 77.51 \\
Legal$\Delta$ & \textbf{83.14} & \textbf{79.68} & \textbf{68.53} & \textbf{62.90} & \textbf{64.80} & \textbf{64.80} & \textbf{75.80} & \textbf{71.38} & \textbf{67.80} & \textbf{64.43} & \textbf{92.94} & 82.19 & \textbf{85.09} & \textbf{78.49} \\
\midrule
\multicolumn{15}{c}{\textit{Qwen2.5-14B-Instruct}} \\
\hdashline
Zero-shot & 84.18 & 79.66 & 58.50 & 52.10 & 42.60 & 55.40 & 76.80 & 64.18 & 74.58 & 65.92 & 90.59 & 86.56 & 87.27 & 80.98 \\
SFT (DiscLaw) & 81.36 & 77.13 & 55.53 & 49.84 & 37.90 & 52.20 & 68.40 & 60.34 & 42.37 & 44.32 & 62.94 & 51.87 & 58.55 & 52.01 \\
DPO & 82.21 & 77.00 & 59.65 & 53.00 & 58.00 & 69.20 & \textbf{79.40} & 68.35 & 67.80 & 59.22 & 91.76 & 85.63 & 84.73 & 77.83 \\
R1-Distill & 84.73 & 80.64 & 59.93 & 53.02 & 45.10 & 69.80 & 78.20 & 67.35 & 72.88 & 66.48 & \textbf{92.35} & 85.31 & 87.27 & 80.86 \\
Legal$\Delta$ & \textbf{85.56} & \textbf{81.10} & \textbf{62.79} & \textbf{56.95} & \textbf{67.10} & \textbf{72.40} & \textbf{79.40} & \textbf{72.19} & \textbf{77.12} & \textbf{68.90} & \textbf{92.35} & \textbf{87.19} & \textbf{88.73} & \textbf{82.86} \\
\bottomrule
\end{tabular}
\caption{Overall Performance.}
\label{tab:lawbench-with-cir3}
\end{table*}
\subsection{Information-Gain Enhanced Legal Reward Modeling}
\label{Information-Gain Enhanced Reward Mechanism}
In this subsection, we introduce an information-theoretic enhancement to reward design. Specifically, we replace the legal reward $R_{\text{legal}}$ component with $R_{\text{info}}$ in Eq.~\ref{eq:reward_raw} for reward modeling:
\begin{equation}
R(o_i) = R_{\text{format}}(o_i) + R_{\text{info}}(a_i).
\end{equation}
The reward $R_{\text{info}}$ incorporates reasoning confidence through differential Q-value analysis to better evaluate the semantic contribution of reasoning processes in legal analysis:
\begin{equation}
R_{\text{info}}(a_i) = R_{\text{legal}}(a_i) \times \sigma\left(\Delta Q \cdot T\right),
\end{equation}
where $\sigma(\cdot)$ is the sigmoid function providing bounded amplification within (0,1) with monotonic enhancement properties, $\Delta Q$ denotes the information quality differential, and $T$ is the temperature parameter that controls the sensitivity of reward modulation for fine-grained control over training dynamics. The multiplicative structure preserves the primacy of correctness while enabling adaptive enhancement based on information content.

Recent work by Li et al.~\cite{li2025generalist} proves that logits from next-token prediction training are mathematically equivalent to Q-functions in offline inverse reinforcement learning: 
\begin{equation}
Q(q, a) = \text{logits}_\theta(a | q),
\end{equation}
where $Q(q, a)$ quantifies the expected value of generating answer $a$ given query $q$, and $\text{logits}_\theta(a | q)$ represents the average unnormalized log-probability scores across all tokens in answer $a$ when generated autoregressively given context $q$. This theoretical foundation enables us to extract Q-values directly from any pre-trained language model and design a differential mechanism to quantify reasoning quality.
For answer $a$, the Q-value gain given by reasoning process $r$ is expressed as:
\begin{equation}
\Delta Q(r) = \text{logits}_\theta(a |q,r) - \text{logits}_\theta(a |q).
\end{equation}
This formulation measures the semantic shift in the model's internal representation space induced by the reasoning process $r$, providing a quantitative measure of how much the reasoning contributes to the model's confidence in generating the correct answer.

From an information-theoretic perspective, given that logits follow the softmax probability distribution:
\begin{equation}
p_i(a) = \frac{e^{\mathrm{logits}_i(a)}}{Z_i},
\end{equation}
where $Z_i = \sum_{a'} e^{\mathrm{logits}_i(a')}$, $a'$ is used to traverse all possible answer tokens. Then $\Delta Q(r)$ can be calculated:
\begin{equation}
\Delta Q(r) = \log \frac{p(a|q,r)}{p(a|q)} + \log \frac{Z_{q,r}}{Z_q}.
\end{equation}
This decomposition reveals the information-theoretic interpretation: \textbf{The first term} $\log \frac{p(a|q,r)}{p(a|q)}$ represents the pointwise mutual information (PMI) between the answer token $a$ and the reasoning process $r$, measuring the correlation between the reasoning chain and the specific answer: a positive value indicates that the reasoning process makes this answer more likely, a negative value means the reasoning actually decreases the probability of this answer, and zero implies no correlation. \textbf{The second term} $\log \frac{Z_{q,r}}{Z_q}$ serves as a global confidence regularizer that captures whether the reasoning process $r$ enhances the model's overall decisiveness: when positive, it indicates that $r$ has sharpened the probability distribution, making model more confident; else, it suggests the reasoning has introduced confusion.



\section{Experimental Methodology}
\label{sec:typestyle}

In this section, we describe the datasets, baselines, evaluation metrics, and implementation details in our experiments. 

\textbf{Datasets}.
We select training data from CAIL2018~\cite{xiao2018cail2018} and JEC\_QA~\cite{zhong2020jec} datasets. 
We first use 450 high-quality reasoning samples distilled from DeepSeek-R1 for foundational reasoning capabilities, then collect 6,981 training and 563 validation samples for GRPO training.
We evaluate on five knowledge-intensive legal reasoning tasks: legal article prediction (SPP-F), criminal charge prediction (CCP), sentencing prediction (SLP), similar case analysis (CAS), and financial amount calculation (CAC). Additionally, we assess generalization on the DiscLaw benchmark~\cite{yue2023disc}.

\textbf{Baselines.}
In our experiments, we compare Legal$\Delta$ with existing law models: Fuzi-7B~\cite{deng2023syllogistic}, HanFei-7B~\cite{HanFei}, LexiLaw-6B~\cite{LexiLaw}, ChatLaw-13B~\cite{cui2023chatlaw}, and DiscLaw-13B~\cite{yue2024lawllm}. These legal models vary in parameter scale and capability, and are trained using diverse methods on a range of legal datasets. To ensure fair and rigorous evaluation, we reimplemented a representative method SFT (DiscLaw) based on the same Qwen2.5 series base models in our experiments. Additionally, we compare with four baseline models: Zero-shot, R1-Distill, DPO, and SFT (DiscLaw) for a comprehensive evaluation.

\textbf{Evaluation Metrics.}
Following recent work~\cite{yue2023disc}, we design comprehensive evaluation metrics: F1 and Jaccard (Jac) scores are used for SPP-F and CCP tasks, while accuracy is used for others.

\textbf{Implementation Details.}
In our experiments, we adopt the Qwen2.5-Instruct series due to its superior reflective capabilities and cognitively informed reasoning behavior~\cite{gandhi2025cognitive}, including the 7B and 14B versions~\cite{qwen2025qwen25technicalreport}, as backbone LLMs to ensure a comprehensive evaluation across different model scales.
During training, we use a batch size of 2 per GPU and a learning rate of 5e-5. For LoRA fine-tuning, we use a rank of 32. We use nucleus sampling with a temperature of 0.9 and generate $G=6$ samples per prompt. To calculate the information gain, we set a temperature of $T=0.2$. 

\section{Evaluation Result}
In this section, we evaluate the performance of various legal LLMs on legal reasoning tasks and then analyze the effectiveness of our information-gain-enhanced reward modeling. 

\subsection{Overall Performance}
This section evaluates the legal reasoning capabilities of Legal$\Delta$ on both in-domain and out-of-domain tasks.

As shown in Table~\ref{tab:lawbench-with-cir3}, Legal$\Delta$ achieves consistently superior performance across different model scales and evaluation settings. On in-domain legal reasoning benchmarks, Legal$\Delta$ outperforms all baselines, with an average improvement of approximately 10\% across Qwen2.5 models of varying capacities. While DPO benefits from preference optimization and R1-Distill leverages high-quality reasoning traces, Legal$\Delta$ outperforms both models, especially on complex reasoning tasks such as SLP and CAS.

Furthermore, the advantages of Legal$\Delta$ extend to out-of-domain scenarios, where it maintains strong performance across diverse evaluation benchmarks. This highlights the generalization capability of our approach, which dynamically optimizes reasoning processes and better captures nuanced legal logic by using the information-gain enhanced reward modeling method.

\subsection{Ablation Study}
To systematically assess the contribution of each core component in the Legal$\Delta$ framework, we conduct ablation experiment on Qwen2.5-14B-Instruct. The results are summarized in Table~\ref{tab:ablation-lawbench}.

Our reinforcement learning based training strategy brings substantial performance gains over the vanilla baseline. Removing the GRPO component (\textit{w/o GRPO}) leads to a marked drop in performance, highlighting the critical role of reinforcement learning in enabling the legal reasoning capability of LLMs. This demonstrates that GRPO's intra-group comparison mechanism effectively exploits quality differences among reasoning trajectories, making it essential for optimizing legal reasoning capabilities. Similarly, the \textit{w/o SFT} setting confirms that both SFT and RL stages are jointly necessary for achieving optimal performance.

The information-gain-enhanced reward mechanism also provides notable performance gains. To verify this, we evaluate a variant that excludes information gain and instead relies solely on traditional reward signals (\textit{w/o Info-Gain}). The resulting performance degradation confirms the important role of incorporating differential Q-value analysis, which enables fine-grained differentiation of reasoning quality. The performance gap between the baseline variant that excludes information gain and our full Legal$\Delta$ model demonstrates that our information-gain enhancement effectively improves legal reasoning ability.
\begin{table}[t]
\centering
\resizebox{\linewidth}{!}{
\small
\begin{tabular}{lcccccc}
\toprule
\multirow{2}{*}{\textbf{Method}} & \textbf{SPP-F} & \textbf{CCP} & \textbf{SLP} & \textbf{CAS} & \textbf{CAC} & \multirow{2}{*}{\textbf{Avg.}} \\
& \textbf{\textit{F1}} & \textbf{\textit{F1}} & \textbf{\textit{Acc}} & \textbf{\textit{Acc}} & \textbf{\textit{Acc}} \\
\midrule

Legal$\Delta$ & 85.56 & 62.79 & \textbf{67.10} & \textbf{72.40} & 79.40 & \textbf{73.45} \\
w/o SFT & \textbf{86.40} & \textbf{66.85} & 57.30 & 71.60 & 77.60 & 71.95 \\
w/o GRPO & 84.73 & 59.93 & 45.10 & 69.80 & 78.20 & 67.55 \\
w/o Info-Gain & 85.09 & 61.69 & 56.20 & 69.60 & \textbf{79.80} & 70.48 \\
\hdashline
Zero-shot & 84.18 & 58.50 & 42.60 & 55.40 & 76.80 & 63.50 \\
\bottomrule
\end{tabular}}
\caption{Ablation Study on Legal Reasoning Tasks.}
\label{tab:ablation-lawbench}
\end{table}
\begin{figure}[t!]
    \centering

    
    \subfigure[Direct Response Scenario.\label{fig:logits_c}]{%
        \includegraphics[width=0.22\textwidth]{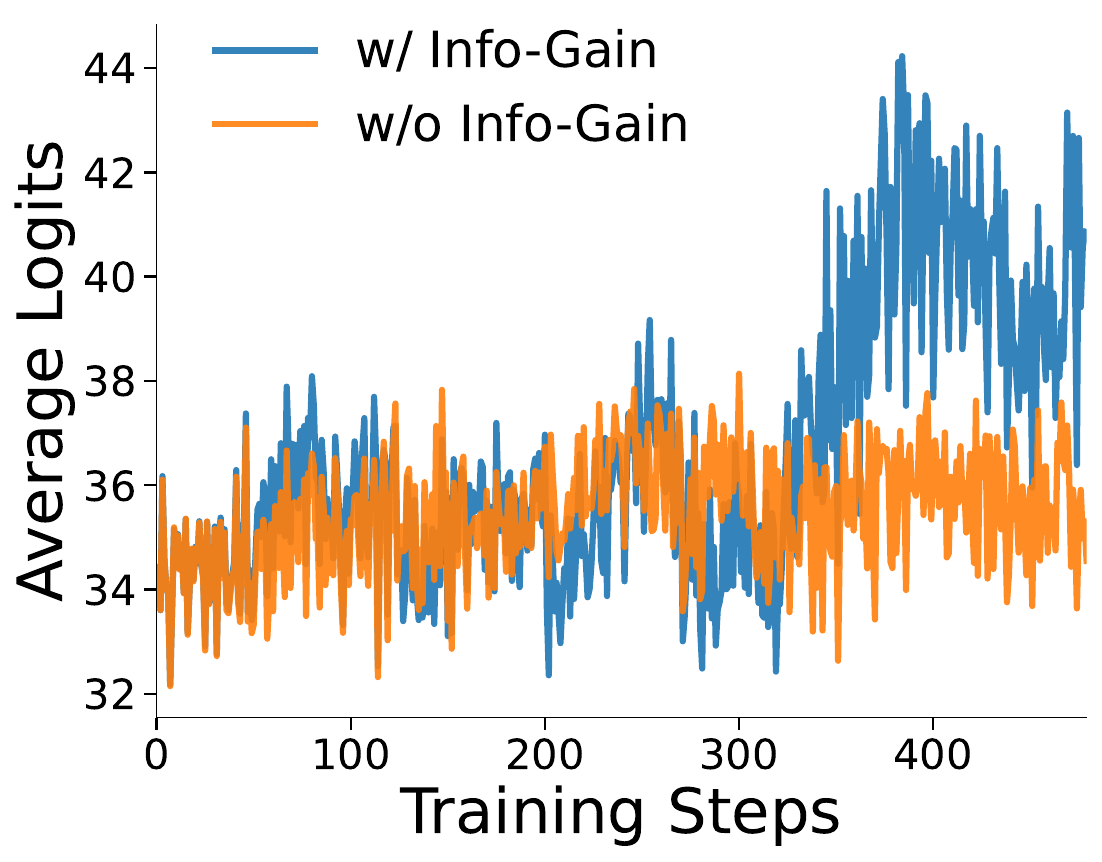}%
    }
    \hfill
    \subfigure[CoT Prompting Scenario.\label{fig:logits_d}]{%
        \includegraphics[width=0.22\textwidth]{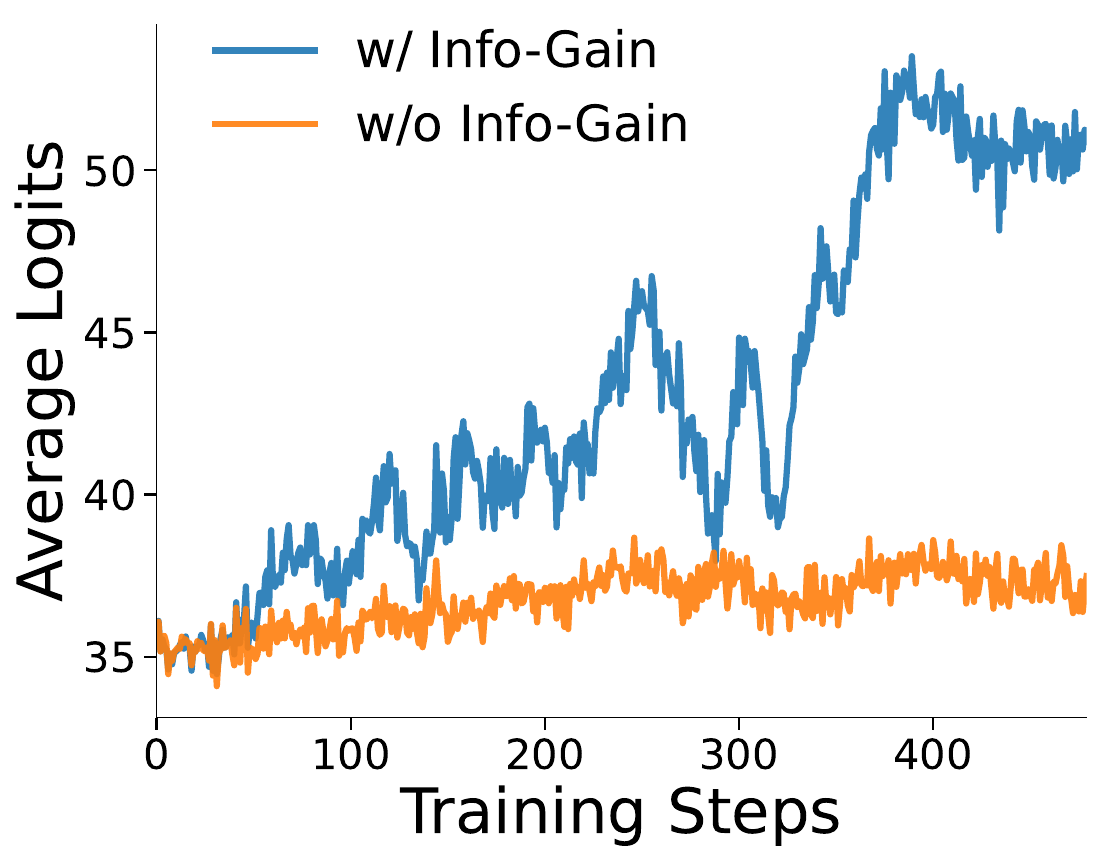}%
    }
       \subfigure[Legal Token Prominence. \label{fig:token_a}]{%
        \includegraphics[width=0.22\textwidth]{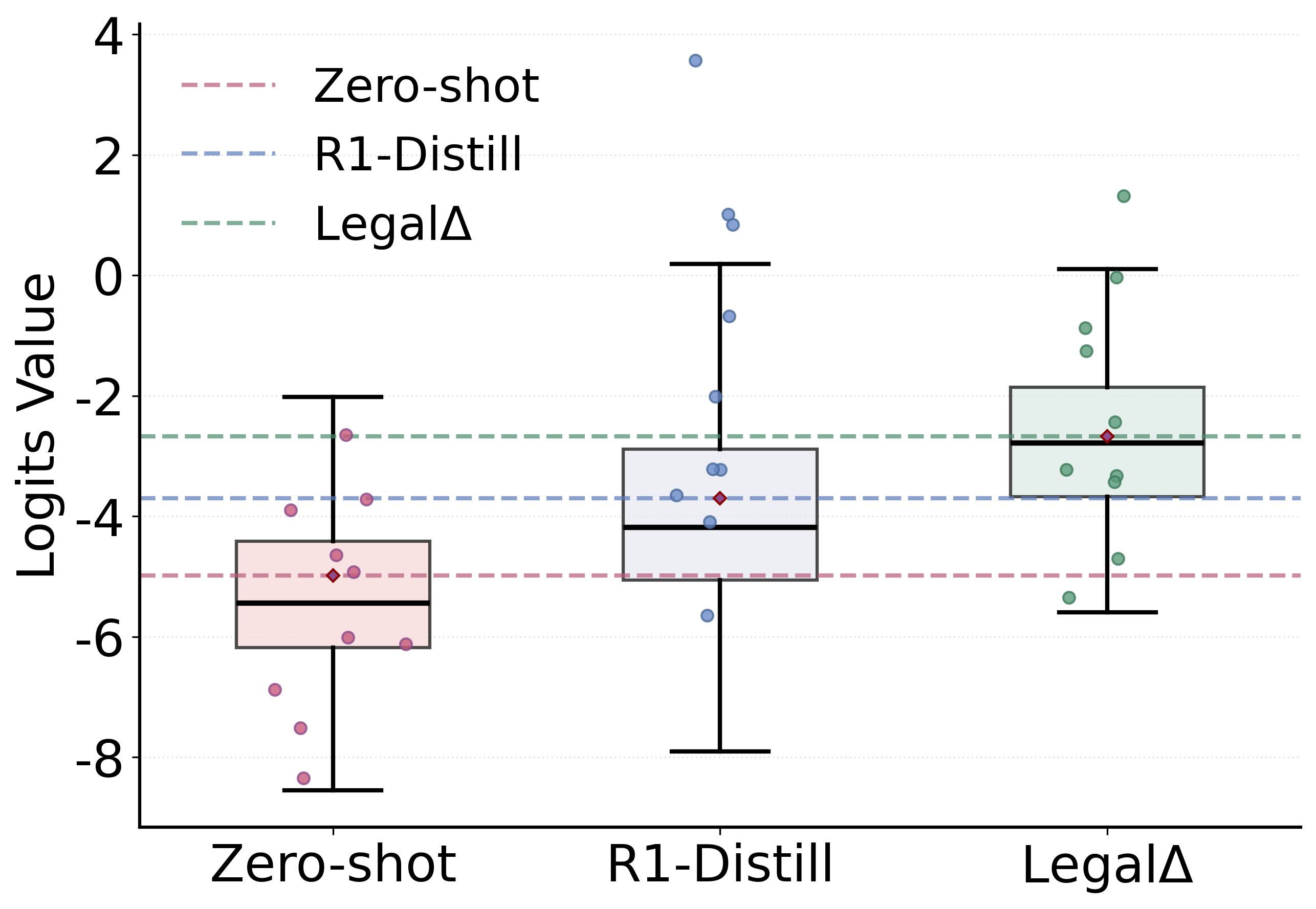}%
    }
    \hfill
    \subfigure[Reasoning Quality. 
    \label{fig:token_b}]{%
        \includegraphics[width=0.22\textwidth]{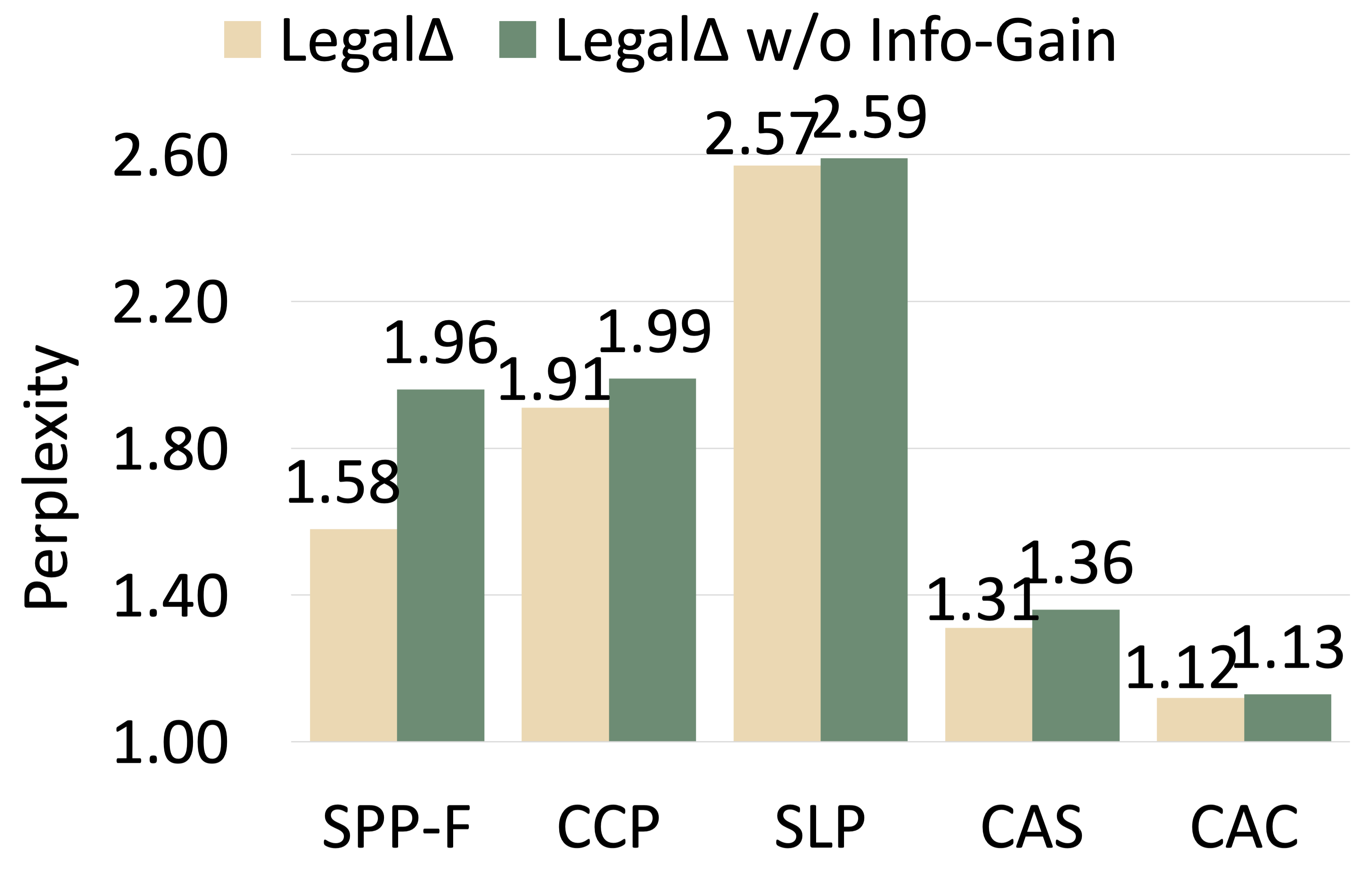}%
    }
    \caption{Performance of Information-Gain Reward Modeling.}
    
    \label{fig:Logits_compare}
\end{figure}


\subsection{Performance of Information-Gain Enhanced Reward}

In this section, we analyze the effectiveness of the information gain reward modeling from two perspectives: (1) its influence on model confidence via logits analysis during training, and (2) its effectiveness on token-level performance quality assessment.

\textbf{Logits Analysis and Model Confidence.}
As demonstrated in Figure~\ref{fig:logits_c} and Figure~\ref{fig:logits_d}, the figures illustrate how average logits change with training steps for the direct response and CoT prompting scenarios, respectively.
To evaluate how the information gain reward modeling contributes to model confidence, we compare the logits performance of two Legal$\Delta$ variants, depending on whether information gain reward is included. 
Logits reflect model confidence in predictions and higher values indicate stronger confidence from more stable internal representations. In direct response scenario, both variants perform similarly initially, but our full Legal$\Delta$ progressively develops higher confidence with widening performance gaps in later training stages. This improvement is more pronounced in CoT prompting scenario, where our Legal$\Delta$ shows substantial and consistent logits gains throughout training while the other variant maintains relatively low confidence levels. These improved logits confirm that our information gain mechanism effectively identifies high-quality reasoning patterns.

\textbf{Performance Quality Assessment.}
As shown in Figure~\ref{fig:token_a}, we evaluate legal token prominence across three different models. Legal token prominence is defined as the difference between logit values for specific legal tokens and the average logits across the sequence, which measures the model's attention to legally relevant content.
We observe progressive improvements: from negative baseline values in Zero-shot, R1-Distill achieves initial improvements through distilled reasoning data, while Legal$\Delta$ further advances through GRPO training with notably higher prominence values and more positive samples, indicating enhanced attention to legal tokens.

We further assess reasoning quality through perplexity analysis by comparing Legal$\Delta$ with and without information gain enhancement in Figure~\ref{fig:token_b}. For each question, we generate reasoning chains and provide them, along with the question, to a vanilla LLM (Qwen2.5-32B-Instruct) to calculate ground truth answer perplexity. Legal$\Delta$ with information gain consistently achieves lower perplexity values across tasks, demonstrating the information-gain reward modeling helps to conduct higher quality reasoning results.


\section{Conclusions}

This paper proposes Legal$\Delta$, a novel framework that encourages LLMs to explore diverse reasoning trajectories and develop robust reasoning strategies for legal tasks by incorporating chain-of-thought–guided information gain into Reinforcement Learning with Verifiable Rewards (RLVR).
Experimental results demonstrate that Legal$\Delta$ effectively balances direct-answer and reasoning-enhanced generation modes, improving model confidence, emphasizing legally relevant tokens and enhancing the quality of the reasoning process compared to baselines.

\section*{Acknowledgments}
This work is partly supported by the National Natural Science Foundation of China (No. 62576082 and No. 62461146205).
\bibliographystyle{IEEEbib}
\bibliography{strings,refs}

\end{document}